\documentclass[conference]{IEEEtran}
\IEEEoverridecommandlockouts
\usepackage{cite}
\usepackage{amsmath,amssymb,amsfonts}
\usepackage{algorithm}
\usepackage{algorithmic}
\usepackage{graphicx}
\usepackage{textcomp}
\usepackage{xcolor}
\usepackage{booktabs}
\usepackage{multirow}
\usepackage{array}
\def\BibTeX{{\rm B\kern-.05em{\sc i\kern-.025em b}\kern-.08em
    T\kern-.1667em\lower.7ex\hbox{E}\kern-.125emX}}

\begin{document}

\title{Improving Asset Allocation in a Fast Moving Consumer Goods B2B Company: An Interpretable Machine Learning Framework for Commercial Cooler Assignment Based on Multi-Tier Growth Targets}  

\author{\IEEEauthorblockN{Renato Castro}
\IEEEauthorblockA{\textit{PUCP Artificial Intelligence Group} \\
\textit{Pontificia Universidad Catolica del Peru}\\
Lima, Peru \\
rcastroc@pucp.edu.pe}
\and
\IEEEauthorblockN{Rodrigo Paredes}
\IEEEauthorblockA{\textit{Department of Industrial Engineering} \\
\textit{Universidad de Lima}\\
Lima, Peru \\
rodrigoparedes@ulima.edu.pe}

\and
\IEEEauthorblockN{Douglas Kahn}
\IEEEauthorblockA{\textit{Department of Economics} \\
\textit{Universidad Nacional del Callao}\\
Lima, Peru \\
dkahn@unac.pe}
}

\maketitle

\begin{abstract}
In the fast-moving consumer goods (FMCG) industry, deciding where to place physical assets, such as commercial beverage coolers, can directly impact revenue growth and execution efficiency. Although churn prediction and demand forecasting have been widely studied in B2B contexts, the use of machine learning to guide asset allocation remains relatively unexplored. This paper presents a framework focused on predicting which beverage clients are most likely to deliver strong returns in volume after receiving a cooler. Using a private dataset from a well-known Central American brewing and beverage company of 3,119 B2B traditional trade channel clients that received a cooler from 2022-01 to 2024-07, and tracking 12 months of sales transactions before and after cooler installation, three growth thresholds were defined: 10\%, 30\% and 50\% growth in sales volume year over year. The analysis compares results of machine learning models such as XGBoost, LightGBM, and CatBoost combined with SHAP for interpretable feature analysis in order to have insights into improving business operations related to cooler allocation; the results show that the best model has AUC scores of 0.857, 0.877, and 0.898 across the thresholds on the validation set. Simulations suggest that this approach can improve ROI because it better selects potential clients to grow at the expected level and increases cost savings by not assigning clients that will not grow, compared to traditional volume-based approaches with substantial business management recommendations.
\end{abstract}


\section{Introduction}

The fast-moving consumer goods (FMCG) industry faces growing pressure to improve how physical assets are allocated to their customers. In the brewing and beverage industry, commercial coolers are one of the most visible and capital intensive assets, directly in trade marketing execution and sales results. When strategically placed in B2B retail environments, these assets can significantly increase sales volume and strengthen customer relationships, making them a key lever for commercial growth.

Despite their strategic importance, cooler deployment decisions are still largely guided by experience, legacy practices, client loyalty, or simple volume-based metrics. These rules often overlook critical signals in customer behavior, historical trends, or the local market context, limiting their effectiveness. As data availability improves and Return on Investment (ROI) expectations increase, there is a need for smarter and more predictive approaches that link asset investments with measurable commercial outcomes. In this context, the ROI expectation is defined by the ratio of sum of the incremental margin and the cost savings of not investing the assets in not potential clients divided by the total coolers investment: purchase, transportation, installation and maintenance.

Although research in B2B customer modeling has focused primarily on churn prevention and retention strategies \cite{gattermann2022, janssens2024}, few studies have addressed the challenge of asset allocation under growth uncertainty. Unlike churn models, which aim to flag customers at risk of leaving, cooler allocation requires identifying clients with high potential to grow when supported by physical investments.

This paper introduces a machine learning framework to predict which B2B beverage clients are most likely to deliver strong sales growth after receiving a commercial cooler. We frame the task as a multithreshold binary classification problem using sales uplift targets of 10\%, 30\%, and 50\%. The models are trained on 3,119 B2B clients using 12 months of pretreatment and posttreatment data and leverage gradient boosting algorithms (XGBoost, LightGBM, CatBoost) with SHapley Additive exPlanations (SHAP) for interpretability. Unlike traditional Customer Life Value estimation, which targets long-term value, our approach focuses on short-term intervention-specific results, making it highly relevant for tactical allocation decisions in B2B operations. Our findings offer a scalable, data-driven method to guide asset deployment, helping decision makers target the right clients, at the right time, with higher confidence and ROI.

The main contributions of this work are:
\begin{itemize}
\item A comprehensive framework for predicting potential customers in B2B FMCG asset allocation scenarios targeting different growth tiers (10\%, 30\%, 50\%) to support flexible business strategies
\item Integration of SHAP-based interpretability to understand feature importance and propose new strategies in the operation
\item Empirical validation on real-world data from 3,119 B2B clients with demonstrated business impact
\end{itemize}

\section{Related Work}

\subsection{B2B Customer Churn Prediction}

In the FMCG research space, Mirković et al. \cite{mirkovic2022} built a churn model using only invoice-level data and still achieved strong performance, showing that robust predictions are possible even with minimal inputs. Their use of a multi-slicing strategy to boost data efficiency, and their finding that ensemble methods consistently outperform single models, are closely aligned with the modeling approach used in this work. Other studies on demand prediction \cite{fmcg2019} and FMCG market analysis \cite{ieee2023fmcg} have shown the promise of ML in the sector, but the majority focus on forecasting or segmentation. The application of machine learning to asset allocation decisions, especially in B2B contexts, is still a gap in the literature that this work aims to address.

\subsection{Interpretable Machine Learning in Business Applications}

The adoption of ML in business-critical applications requires model interpretability to build stakeholder trust and enable actionable insights. Daz et al. \cite{diaz2022} applied explainable AI techniques to the prediction of churn of B2B customers for enterprise software, using SHAP values to reveal the most influential factors behind the risk of churn.

Recent advances in interpretable ML have shown particular promise in customer prediction tasks. Maan et al. \cite{maan2023} developed transparent churn prediction models using XGBoost with SHAP values, while telecom research has achieved high AUC (0.79) with XAI-Churn TriBoost while maintaining interpretability \cite{telecom2025}. The integration of SHAP with gradient boost models has proven effective in identifying key drivers of customer behavior \cite{churn2024}, enabling targeted strategies based on the importance of the features \cite{telecom2024}.

\subsection{Resource Allocation and Asset Optimization}

Although there is extensive research on customer prediction and churn management, the application of machine learning to physical asset allocation in B2B contexts remains limited. Wang et al. \cite{wang2017} proposed a general framework for resource allocation using historical scenario data, demonstrating the superiority of machine learning over conventional optimization methods.

Colias et al. \cite{colias2023} addressed the optimization of B2B products using mixed logit models and non-linear programming, integrating machine learning with economic theory to optimize features and pricing. However, the specific challenge of predicting ROI for physical asset deployment in FMCG B2B contexts represents a gap in the literature. Our work addresses this gap by combining profit-driven modeling approaches with interpretable ML techniques to optimize asset allocation decisions.

\section{Problem Formulation}

\subsection{Business Context}

The allocation decision involves selecting which B2B clients, such as retailers, restaurants, and so on, should receive these assets to maximize overall business impact. Given limited resources and high asset costs, a commercial cooler with all related logistics costs is roughly \$974. Optimal allocation is crucial to direct these assets to clients with attributes that support sales growth, while avoiding placement with clients unlikely to show significant growth.

\subsection{Dataset Description}

\begin{figure}[htbp]
    \centering
    \includegraphics[width=\linewidth]{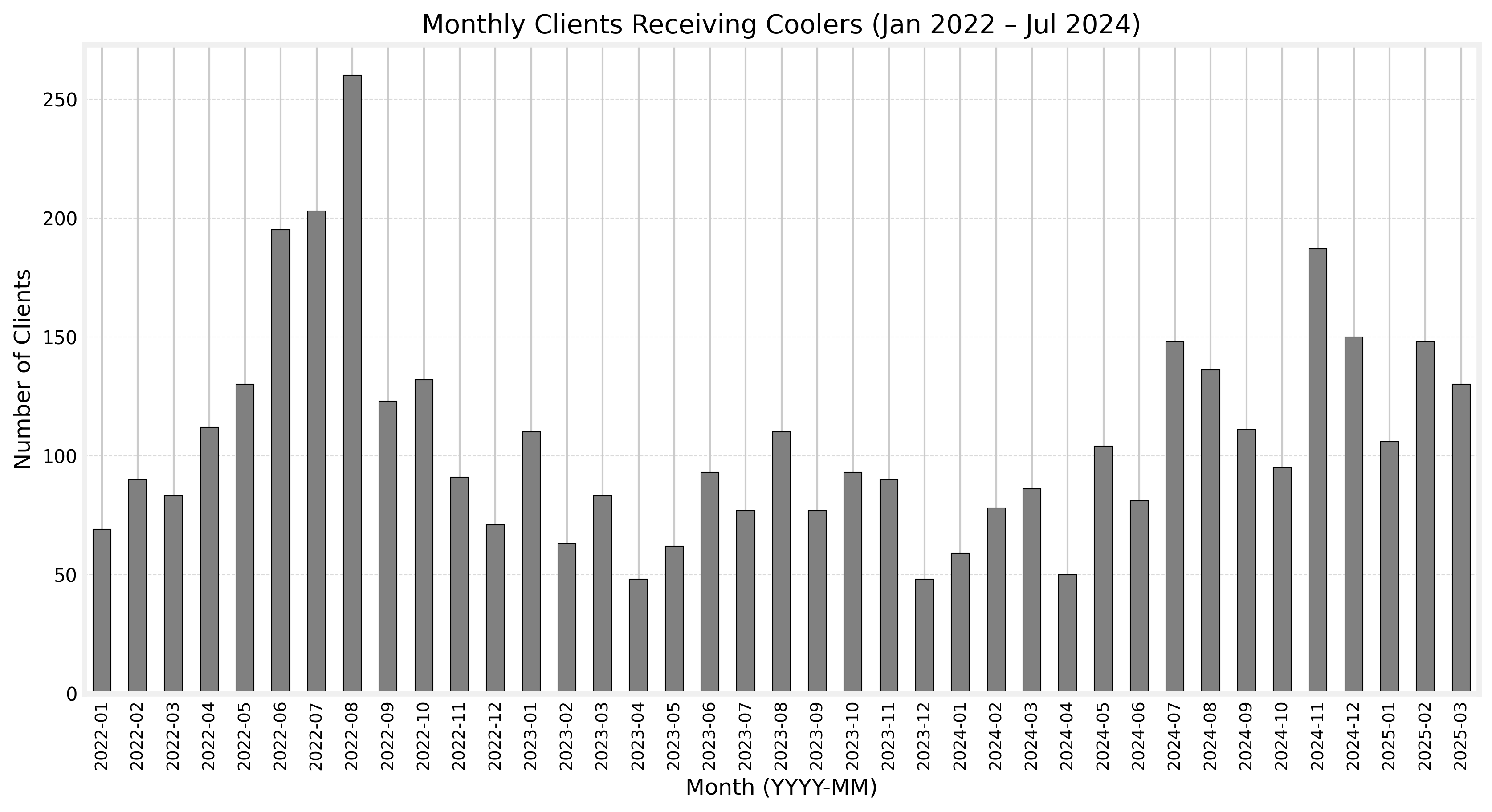}
    \caption{Monthly number of unique B2B clients receiving coolers from January 2022 to March 2025.}
    \label{fig:monthly_clients}
\end{figure}

Our dataset comprises transactional data from 3,119 B2B clients spanning January 2022 to March 2025 (see Fig.~\ref{fig:monthly_clients}) Each client record includes:

\begin{itemize}
\item \textbf{Historical sales transactions}: Monthly volume (liters), revenue, number of products, product lines share
\item \textbf{Behavioral features}: Order frequency, payment history, and purchase patterns
\item \textbf{Market context}: Country census data for 2018, regional indicators, and competition density
\end{itemize}

\subsection{Target Variable Definition}

In this work, we define sales volume as the total number of hectoliters sold by the company to each B2B client, aggregated across all beverage product lines (beer, water, soft drinks, etc.). This unit serves as the standardized metric for evaluating client growth, both in historical patterns and post-allocation uplift.

We define success through sales volume growth metrics, creating three binary classification targets:

\begin{equation}
y_i^{(k)} = \begin{cases}
1 & \text{if } \frac{V_{post,i} - V_{pre,i}}{V_{pre,i}} \geq \tau_k \\
0 & \text{otherwise}
\end{cases}
\end{equation}

Equation (1) shows variables called $V_{pre,i}$ and $V_{post,i}$ that represent cumulative volumes 12 months before and after treatment for the client $i$, and $\tau_k \in \{10\%, 30\%, 50\%\}$ represent growth thresholds. These thresholds were defined as an inner business objective of the company in order to simulate scenarios according to business strategies.

This growth-based framework is fundamentally different from churn prediction approaches \cite{mirkovic2022, diaz2022}, which generally rely on binary labels to identify clients at risk of leaving. Instead of focusing on retention, our objective is to identify clients who are likely to grow their sales volume after asset placement or any treatment a business is looking forward to achieve.

It is also important to clarify that this research does not attempt to prove that the cooler itself is the unique driver of growth. The models are trained only on historical data, without any causal assumptions. This research points to capture the conditions and client profiles historically associated with volume uplift after cooler assignment. That growth may result from multiple factors: commercial attention, market dynamics, product strategy, or a combination of interventions, of which the cooler is just one. In summary, the purpose of the present research is to support better targeting decisions, not to isolate or attribute causality to a single driver.

\subsection{Evaluation Framework}

The multi-threshold approach enables flexible business strategies:
\begin{itemize}
\item \textbf{10\% threshold}: Broad allocation strategy maximizing reach
\item \textbf{30\% threshold}: Balanced approach targeting moderate-to-high performers
\item \textbf{50\% threshold}: Focused strategy for highest-impact opportunities
\end{itemize}

Although uplift modeling approaches \cite{gubela2021} could estimate individual treatment effects, our observational data lack a proper control group, as cooler allocation was not randomized. Instead, we focus on identifying customers with high growth potential post-treatment, acknowledging that some growth may be organic rather than treatment-induced. Future work with randomized allocation could enable estimation of causal effects.

The multithreshold design provides flexibility for different business scenarios: during expansion phases, companies might target the 10\% threshold to maximize market coverage; during efficiency drives, the 50\% threshold ensures resources go only to highest impact opportunities. This adaptability makes the framework valuable for varying business cycles and strategic priorities.

\section{Methodology}\subsection{Data Preprocessing and Feature Engineering}

The proposed preprocessing pipeline addresses common data quality issues in transactional datasets and features that capture both behavioral and contextual client characteristics:

\begin{enumerate}

\item \textbf{Temporal alignment}: All client histories are aligned relative to the cooler installation month to ensure consistency in pre- and post-treatment windows. Depending on the month of cooler delivery, the previous 12-month window was studied in order to prepare the feature engineering.

\item \textbf{Missing value handling}: We do not perform explicit imputation, as gradient boosting models such as XGBoost, LightGBM, and CatBoost can natively handle missing values during training. This allows the model to learn optimal splitting rules when encountering nulls, preserving the original data distribution without introducing imputation bias. However, for a business operation, this identification can help fix the retrieval of these missing values.

\item \textbf{Outlier treatment}: Numerical outliers are capped at the upper bound defined by $Q3 + 1.5 \times IQR$, where $Q3$ is the 75th percentile and $IQR$ is the interquartile range.

\item \textbf{Feature engineering}: This step results in a dataset of 3,119 client observations and 3,469 variables.
    \begin{itemize}
    \item Rolling volumes and revenues over 3, 6, and 12 month windows, calculated globally and by product line and brand. These include sum, mean, max, min, and standard deviation.
    
    \item Recency and frequency metrics derived from sales activity aggregated similarly by 3, 6, and 12 month windows and segmented by product line and brand. Recency is defined as the number of days between purchases and frequency, defined as the distinct count of purchase days per month.
    
    \item Rolling statistics of discounts aggregated similarly by 3, 6, and 12 month windows and product grouping. These include sum, mean, max, min, and standard deviation.
    
    \item Sociodemographic variables constructed by associating the geographic coordinates of each client (latitude and longitude) with the nearest census polygon. These polygons are defined by official national boundaries and contain attributes such as average household income, population density, and educational attainment. A spatial join was performed to enrich each client with the corresponding census features of the matched polygon.
    \end{itemize}
    
\item \textbf{Feature filtering}: Before model training, redundancy and leakage risk characteristics were tested. Variables with extremely high correlation to each other and to the target (Pearson correlation $|r| > 0.80$) were removed to reduce multicollinearity and prevent leakage and overly optimistic evaluation. This step results in a dataset of 3,119 client observations and 574 variables.

\end{enumerate}

Unlike churn-focused models that emphasize declining engagement patterns, our feature set emphasizes positive growth trends and structural constraints. The underlying hypothesis is that clients that show organic sales momentum - but potentially limited by infrastructure (e.g. lack of cooling capacity) - are the best candidates for cooler allocation. As such, many features are designed to capture latent demand, operational friction, and commercial readiness.

\subsection{Class Imbalance Handling}

\begin{table}[htbp]
\centering
\caption{Class Distribution Across Target Growth Thresholds}
\label{tab:class_distribution}
\begin{tabular}{lcc}
\hline
\textbf{Target Variable} & \textbf{Class 0 (\%)} & \textbf{Class 1 (\%)} \\
\hline
growth target 10\% & 53.99 & 46.00 \\
growth target 30\% & 58.15 & 41.84 \\
growth target 50\% & 62.05 & 37.95 \\
\hline
\end{tabular}
\end{table}

Each binary classification task in this study corresponds to a specific sales growth threshold: 10\%, 30\%, and 50\%. As shown in Table~\ref{tab:class_distribution}, the proportion of positive cases (clients who achieved the growth target) decreases as the threshold increases. Although target 10 is relatively balanced, targets 30 and 50 exhibit an increasing class imbalance.

To address this, stratified sampling was used in both the train-validation split and the cross-validation process to preserve class ratios throughout model development. Alternative approaches such as random oversampling, undersampling, and SMOTE are research techniques to be applied when there is an imbalanced target. However, given the relatively mild imbalance levels and the performance of tree-based models with internal weighting, we prioritized the native capabilities of each algorithm for final training. This ensured the robustness of the model without introducing synthetic data or altering the underlying distribution.

\subsection{Machine Learning Model Solutions}

We evaluated three gradient-boosting implementations known for superior performance in tabular data:

\begin{itemize}

\item \textbf{XGBoost}: Tree-based boosting algorithm with regularization and shrinkage, optimized for speed and accuracy on structured data.
\item \textbf{LightGBM}: Gradient boosting framework using histogram-based tree learning and leaf-wise growth to improve speed and reduce memory usage.
\item \textbf{CatBoost}: Gradient boosting method with efficient handling of categorical variables and ordered boosting to reduce overfitting.

\end{itemize}

These algorithms excel at capturing non-linear relationships between features and can naturally handle the mix of numerical and categorical variables typical in B2B datasets.

\subsection{Training and Evaluation Strategy}

To ensure unbiased performance estimation and simulate real-world deployment, we separate the data into two clearly defined stages: a training set (used for model development and validation via cross-validation) and a holdout test set (retained entirely for the final evaluation to ensure generalizability). Training and holdout test sets are split 80\% and 20\%, respectively, using a stratified approach to preserve target distributions.

\textbf{Training Strategy:} 

Our training approach ensures robust model evaluation and model interpretability, combining cross-validation, hyperparameter tuning, and iterative feature selection.

\begin{enumerate}
    \item \textbf{Cross-Validation}: We apply 5-fold stratified cross-validation in the training set to produce stable performance estimates and reduce variance.
    \item \textbf{Hyperparameter Optimization}: We use Optuna for hyperparameter tuning across all evaluated models, maximizing validation AUC while maintaining computational efficiency.
    \item \textbf{SHAP-Based Feature Selection}: We implement iterative feature elimination using SHAP values to improve interpretability and reduce model complexity.
    
\end{enumerate}
\begin{algorithm}[htbp]
\caption{Cross-Validated Training Pipeline with Hyperparameter Optimization and SHAP-based Feature Selection}
\label{alg:training_pipeline}
\begin{algorithmic}[1]
\STATE \textbf{Input:} Training Set
\STATE Initialize feature set $F = $ all features
\STATE \textbf{Repeat until convergence or no improvement in CV performance:}
    \STATE \quad Split the training set using stratified 5-fold CV

    \STATE \quad \textbf{Stage 1: Hyperparameter Optimization}
    \STATE \quad Initialize search space for hyperparameters
    \STATE \quad \textbf{For} each Optuna trial \textbf{do}
        \STATE \quad \quad Sample hyperparameters
        \STATE \quad \quad Evaluate model using 5-fold CV on current $F$
        \STATE \quad \quad Store trial performance
    \STATE \quad \textbf{End For}
    \STATE \quad Select best hyperparameters from trials

    \STATE \quad \textbf{Stage 2: SHAP-based Feature Selection}
    \STATE \quad \textbf{While} CV performance improves
        \STATE \quad \quad Train model on current $F$ using best hyperparameters
        \STATE \quad \quad Compute SHAP values on validation folds
        \STATE \quad \quad Rank features by mean($|\text{SHAP}|$)
        \STATE \quad \quad Remove bottom 10\% least important features
        \STATE \quad \quad $F \leftarrow$ remaining features
        \STATE \quad \quad Evaluate model using 5-fold CV on $F$
    \STATE \quad \textbf{End While}
\STATE \textbf{Until no further gain in overall CV performance}
\STATE Output: Final model and selected feature set $F$
\end{algorithmic}
\end{algorithm}

This approach ensures model interpretability while preserving predictive performance. Following the guidance of explainable AI research in B2B decision contexts \cite{diaz2022}, we prioritize features that are both statistically strong and operationally actionable. Non-actionable features, such as static geographic codes, are down-weighted in favor of commercial and behavioral variables that drive allocation decisions.

\begin{table*}[htbp]
\caption{Model Performance Comparison Across Growth Targets (Holdout Set)}
\centering
\begin{tabular}{lcccccc}
\toprule
\textbf{Growth Target} & \textbf{Model} & \textbf{AUC (CV)} & \textbf{Precision (CV)} & \textbf{AUC (Test)} & \textbf{Precision (Test)} & \textbf{Recommended} \\
\midrule
\multirow{3}{*}{10\%} 
& XGBoost    & 0.850 & 0.772 & 0.860 & \textbf{0.798} & Consider \\
& LightGBM   & \textbf{0.855} & 0.772 & 0.857 & 0.783 & Consider \\
& CatBoost   & 0.848 & \textbf{0.774} & \textbf{0.862} & 0.796 & \textbf{Selected} \\
\midrule
\multirow{3}{*}{30\%} 
& XGBoost    & \textbf{0.872} & 0.786 & 0.872 & 0.782 &  \\
& LightGBM   & 0.869 & \textbf{0.790} & \textbf{0.877} & 0.781 & \textbf{Selected} \\
& CatBoost   & 0.865 & 0.789 & 0.866 & \textbf{0.783} & Alternative \\
\midrule
\multirow{3}{*}{50\%} 
& XGBoost    & \textbf{0.869} & 0.769 & 0.894 & 0.796 &  \\
& LightGBM   & 0.865 & 0.766 & \textbf{0.898} & \textbf{0.802} & \textbf{Selected} \\
& CatBoost   & 0.861 & \textbf{0.771} & 0.891 & 0.793 & Consider \\
\bottomrule
\end{tabular}
\label{tab:multi_threshold_results}
\end{table*}

\textbf{Evaluation Strategy:} The selected models -after optimization and feature pruning - are retrained throughout the training set using the best hyperparameter configuration. We then predict on the holdout set and compute key metrics: AUC and Precision. These metrics are chosen to reflect both statistical performance and business applicability.

By decoupling model development from final evaluation, we avoid overfitting and ensure a rigorous, real-world estimation of predictive power across the three growth thresholds (10\%, 30\%, 50\%).

\section{Results}

\subsection{Model Performance Across Thresholds}

Table~\ref{tab:multi_threshold_results} summarizes model performance across the three classification targets: 10\%, 30\%, and 50\% volume growth. Metrics reported include AUC and Precision (at a threshold of 0.5) for both cross-validation (CV) and holdout evaluation (Test).

For the 10\% target, CatBoost achieved the highest AUC on the holdout set (0.862), while XGBoost obtained the highest precision (0.798). However, CatBoost provided the best trade-off between discriminative ability and reliability, making it the preferred model for broad cooler allocation scenarios.

At the 30\% threshold, LightGBM delivered the top AUC in both validation (0.869) and test (0.877), although CatBoost narrowly outperformed in test set precision. Both models are strong candidates depending on the strategic focus — whether ranking performance (LightGBM) or exact targeting (CatBoost) is prioritized.

For the 50\% target, LightGBM emerged as the clear winner with the highest AUC (0.898) and precision (0.802) on the holdout set, indicating its strong potential for high-ROI deployments where selectivity is critical.

These results highlight the importance of tuning model selection to the business objective. While all three models demonstrate robust performance, LightGBM consistently delivers top-tier accuracy and is especially suited for high-growth targeting in resource-constrained environments.

\begin{table*}[htbp]
\caption{Detailed Performance Metrics for Target 30\% Growth (Holdout Test Set)}
\centering
\begin{tabular}{lcccccc}
\toprule
\textbf{Model} & \textbf{AUC} & \textbf{F1-Score} & \textbf{Precision (0.5)} & \textbf{Recall (0.5)} & \textbf{Precision@100} & \textbf{Precision@500} \\
\midrule
XGBoost   & 0.872 & 0.748 & 0.782 & 0.716 & 0.78 & 0.562 \\
LightGBM  & \textbf{0.877} & \textbf{0.750} & 0.781 & \textbf{0.722} & 0.73 & \textbf{0.584} \\
CatBoost  & 0.866 & 0.745 & \textbf{0.783} & 0.711 & \textbf{0.79} & 0.570 \\
\bottomrule
\end{tabular}
\label{tab:target30_detailed}
\end{table*}

\subsection{Results for 30\% Growth Target (Holdout Test Set)}

Table~\ref{tab:target30_detailed} presents detailed evaluation metrics for the growth target 30\%, comparing XGBoost, LightGBM, and CatBoost using standard classification and ranking-based metrics. 

LightGBM achieved the highest AUC (0.877), F1-Score (0.750), and Precision@500 (0.584), indicating strong overall discriminative performance and reliable high-volume targeting. These results suggest that LightGBM is the most balanced model for mid-tier growth prediction, especially when asset deployment involves targeting several hundred clients.

CatBoost achieved the highest Precision@100 (0.790), making it particularly effective when only a small number of coolers are available and the allocation must be extremely selective. It also maintained the highest Precision at threshold 0.5 (0.783), highlighting its value in low-recall, high-precision contexts where misallocation is costly.

From a business perspective, the choice between LightGBM and CatBoost depends on the resource availability: 
\begin{itemize}
    \item If coolers are limited and precision at the very top matters most (e.g., targeting top 100 clients), CatBoost provides the most confident targeting.
    \item If the strategy involves a broader deployment with moderate selectivity, LightGBM offers a better balance between identifying opportunities and costs savings of incorrect asset allocation.
\end{itemize}

\subsection{Feature Importance Analysis and Interpretability}

\begin{figure}[htbp]
\centering
\includegraphics[width=0.47\textwidth]{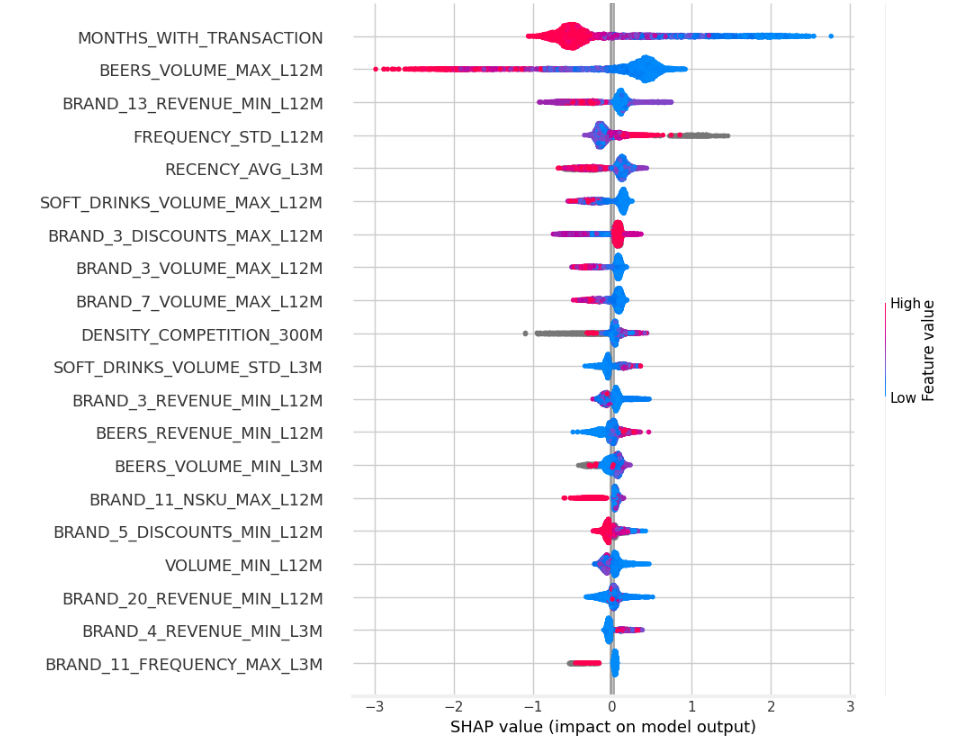}
\caption{SHAP summary plot for 30\% growth model showing top 20 features. Color indicates feature value; x-axis indicates SHAP impact on prediction.}
\label{fig:shap_summary}
\end{figure}

To better understand the factors driving model predictions for the 30\% growth target, we analyzed SHAP values across the top 20 features. Figure~\ref{fig:shap_summary} displays the SHAP summary plot, revealing how each feature influences the output across the validation set.

The most influential variable was \textbf{MONTHS WITH TRANSACTION}, indicating that clients with few month purchases are more likely to experience significant growth post-cooler allocation compared to the consistent clients, the asset allocation could activate or build a better relationship of loyalty with the client. High values of this feature (shown in red) consistently push the prediction toward the positive class.

Another strong driver was \textbf{BEERS VOLUME MAX L12M}, which captures a peak sales of a client in the beer category over the past year. High beer volumes correlate with higher SHAP contributions, validating that high-capacity clients offer greater potential uplift. In contrast, \textbf{BRAND 13 REVENUE MIN L12M}, which measures minimal revenue for a given brand, suggests that clients with low historical performance in specific brands may still represent growth opportunities when properly targeted.

Temporal features like \textbf{FREQUENCY STD L12M} and \textbf{RECENCY AVG L3M} capture behavioral regularity and recent activity, respectively.  SHAP values show that stable ordering patterns and recent participation contribute positively to the prediction of the model, signaling readiness for scaled intervention.

Finally, competitive density — captured by \textbf{DENSITY COMPETITION 300M} — exhibits a negative effect: clients in high-competition areas are less likely to benefit from cooler allocation, potentially due to lower pricing power or market saturation.

\section{Conclusion}

This study presents a machine learning framework to optimize the allocation of commercial coolers in a brewery and beverage B2B FMCG company. By modeling sales growth as a multi-threshold classification problem, we demonstrate how interpretable algorithms like XGBoost, LightGBM, and CatBoost can effectively predict client-level ROI post-allocation. Across all thresholds, the models achieved strong AUC and precision scores on unseen test data. LightGBM stood out as the most robust model, achieving AUC scores of 0.857, 0.877, and 0.898 across the three growth thresholds in the validation set. The proposed methodology offers a scalable and business-aligned decision support system that balances reach, selectivity, and cost-efficiency in asset deployment.

\section{Acknowledgment}

The authors acknowledge the use of artificial intelligence (AI) tools in the preparation of this manuscript, in accordance with IEEE’s guidelines on AI-generated content.
  
The authors used OpenAI ChatGPT (GPT-4o) and Anthropic Claude Opus 4 to assist with grammar polishing, formatting, and language refinement in Abstract, Introduction, Methodology, and Results. All scientific concepts, analysis, methodology workflow and conclusions are the sole responsibility of the authors.

The authors confirm that the use of AI did not compromise the scientific integrity of the paper.

\bibliographystyle{IEEEtran}
\bibliography{references}

@article{gattermann2022,
author = {Gattermann-Itschert, Theresa and Thonemann, Ulrich},
year = {2022},
month = {11},
pages = {134-147},
title = {Proactive customer retention management in a non-contractual {B2B} setting based on churn prediction with random forests},
volume = {107},
journal = {Industrial Marketing Management},
doi = {10.1016/j.indmarman.2022.09.023}
}

@article{janssens2024,
author = {Janssens, Bram and Bogaert, Matthias and Bagué, Astrid and Van den Poel, Dirk},
year = {2022},
month = {03},
pages = {267-293},
title = {{B2B}oost: instance-dependent profit-driven modelling of {B2B} churn},
volume = {341},
journal = {Annals of Operations Research},
doi = {10.1007/s10479-022-04631-5}
}

@article{mirkovic2022,
author = {Mirkovic, Milan and Vučković, Teodora and Stefanović, Darko and Anderla, Andras and Gracanin, Danijela},
year = {2022},
month = {05},
pages = {5001},
title = {Customer Churn Prediction in {B2B} Non-Contractual Business Settings Using Invoice Data},
volume = {12},
journal = {Applied Sciences},
doi = {10.3390/app12105001}
}

@article{diaz2022,
author = {Marín Díaz, Gabriel and Galan Hernandez, Jose Javier and Carrasco, Ramón},
year = {2022},
month = {10},
pages = {3896},
title = {{XAI} for Churn Prediction in {B2B} Models: A Use Case in an Enterprise Software Company},
volume = {10},
journal = {Mathematics},
doi = {10.3390/math10203896}
}

@misc{maan2023,
      title={Customer Churn Prediction Model using Explainable Machine Learning}, 
      author={Jitendra Maan and Harsh Maan},
      year={2023},
      eprint={2303.00960},
      archivePrefix={arXiv},
      primaryClass={cs.LG},
}

@article{telecom2025,
title = {A data-driven approach with explainable artificial intelligence for customer churn prediction in the telecommunications industry},
journal = {Results in Engineering},
volume = {26},
pages = {104629},
year = {2025},
issn = {2590-1230},
doi = {https://doi.org/10.1016/j.rineng.2025.104629},
author = {Daniyal Asif and Muhammad Shoaib Arif and Aiman Mukheimer}
}

@article{churn2024,
title = {Explaining customer churn prediction in telecom industry using tabular machine learning models},
journal = {Machine Learning with Applications},
volume = {17},
pages = {100567},
year = {2024},
issn = {2666-8270},
doi = {https://doi.org/10.1016/j.mlwa.2024.100567},
author = {Sumana Sharma Poudel and Suresh Pokharel and Mohan Timilsina}
}

@article{telecom2024,
author = {Chang, Victor and Hall, Karl and Xu, Qianwen and Amao, Folakemi and Ganatra, Meghana Ashok and Benson, Vladlena},
year = {2024},
month = {05},
pages = {231},
title = {Prediction of Customer Churn Behavior in the Telecommunication Industry Using Machine Learning Models},
volume = {17},
journal = {Algorithms},
doi = {10.3390/a17060231}
}

@article{fmcg2019,
title = {Machine Learning in Predicting Demand for Fast-Moving Consumer Goods: An Exploratory Research},
journal = {IFAC-PapersOnLine},
volume = {52},
number = {13},
pages = {737-742},
year = {2019},
note = {9th IFAC Conference on Manufacturing Modelling, Management and Control MIM 2019},
issn = {2405-8963},
doi = {https://doi.org/10.1016/j.ifacol.2019.11.203},
author = {Elcio Tarallo and Getúlio K. Akabane and Camilo I. Shimabukuro and Jose Mello and Douglas Amancio}
}

@INPROCEEDINGS{ieee2023fmcg,
  author={Kumari, C. Shyamala and Deepu, B S V Vidya and G, Dheeraj and Surampudi, Vinay Kumar Sai},
  booktitle={2023 3rd International Conference on Pervasive Computing and Social Networking (ICPCSN)}, 
  title={{FMCG} Market Analysis for Wholesalers and Retailers Using Machine Learning}, 
  year={2023},
  volume={},
  number={},
  pages={323-328},
  doi={10.1109/ICPCSN58827.2023.00059}}

@ARTICLE{wang2017,
  author={Wang, Jun-Bo and Wang, Junyuan and Wu, Yongpeng and Wang, Jin-Yuan and Zhu, Huiling and Lin, Min and Wang, Jiangzhou},
  journal={IEEE Network}, 
  title={A Machine Learning Framework for Resource Allocation Assisted by Cloud Computing}, 
  year={2018},
  volume={32},
  number={2},
  pages={144-151},
  doi={10.1109/MNET.2018.1700293}}

@unknown{colias2023,
author = {Colias, John and Park, Stella and Horn, Elizabeth},
year = {2023},
month = {08},
pages = {},
title = {Optimizing B2B Product Offers with Machine Learning, Mixed Logit, and Nonlinear Programming},
doi = {10.48550/arXiv.2308.07830}
}

@article{gubela2021,
title = {Uplift modeling with value-driven evaluation metrics},
journal = {Decision Support Systems},
volume = {150},
pages = {113648},
year = {2021},
note = {Interpretable Data Science For Decision Making},
issn = {0167-9236},
doi = {https://doi.org/10.1016/j.dss.2021.113648},
author = {Robin M. Gubela and Stefan Lessmann}
}

\end{document}